\documentclass[sigconf]{acmart}
\pdfoutput=1
\usepackage{amsmath}
\usepackage{wrapfig}
\usepackage{graphicx}
\graphicspath{ {./} }

\DeclareMathOperator*{\argmin}{arg\,min}
\AtBeginDocument{%
  \providecommand\BibTeX{{%
    \normalfont B\kern-0.5em{\scshape i\kern-0.25em b}\kern-0.8em\TeX}}}

\copyrightyear{2022}
\acmYear{2022}
\setcopyright{rightsretained}
\acmConference[FAccT '22]{2022 ACM Conference on Fairness, Accountability, and Transparency}{June 21--24, 2022}{Seoul, Republic of Korea}
\acmBooktitle{2022 ACM Conference on Fairness, Accountability, and Transparency (FAccT '22), June 21--24, 2022, Seoul, Republic of Korea}\acmDOI{10.1145/3531146.3533161}
\acmISBN{978-1-4503-9352-2/22/06}

\acmConference[FAccT '22]{FAccT '22}{June 21--24, 2022}{Seoul, South Korea}
\acmBooktitle{FAccT '22, June 21--24, 2022, Seoul, South Korea}

\begin{document}

\title[Subverting machines, fluctuating identities: Re-learning human categorization]{Subverting machines, fluctuating identities: \\Re-learning human categorization}


\author{Christina Lu}
\authornote{Both authors contributed equally to this research.}
\email{christinalu@deepmind.com}
\orcid{0000-0002-3969-8923}
\affiliation{%
  \institution{DeepMind}
  \city{London}
  \country{UK}
}

\author{Jackie Kay}
\authornotemark[1]
\email{kayj@deepmind.com}
\orcid{0000-0001-9593-695X}
\affiliation{%
  \institution{DeepMind}
  \city{London}
  \country{UK}
}

\author{Kevin R. McKee}
\email{kevinrmckee@deepmind.com}
\orcid{0000-0002-4412-1686}
\affiliation{%
  \institution{DeepMind}
  \city{London}
  \country{UK}
}

\renewcommand{\shortauthors}{Lu and Kay, et al.}

\begin{abstract}
Most machine learning systems that interact with humans construct some notion of a person’s “identity,” yet the default paradigm in AI research envisions identity with essential attributes that are discrete and static. In stark contrast, strands of thought within critical theory present a conception of identity as malleable and constructed entirely through interaction; a \textit{doing} rather than a \textit{being}. In this work, we distill some of these ideas for machine learning practitioners and introduce a theory of identity as autopoiesis, circular processes of formation and function. We argue that the default paradigm of identity used by the field immobilizes existing identity categories and the power differentials that co-occur, due to the absence of iterative feedback to our models. This includes a critique of emergent AI fairness practices that continue to impose the default paradigm. Finally, we apply our theory to sketch approaches to autopoietic identity through multilevel optimization and relational learning. While these ideas raise many open questions, we imagine the possibilities of machines that are capable of expressing human identity as a relationship perpetually in flux.
\end{abstract}

\keywords{algorithmic fairness,
theories of identity,
social construction,
identity systems}

\maketitle

\section{Introduction} 
\label{intro}

Interact with the vast majority of technological systems today, and the underlying artificial intelligence model constructs a representation of you. The usage of such representations ranges from relatively benign (recommender systems optimized to serve you the best ad, video, song or search result) to more nefarious (aiding federal judges by calculating recidivism rates, targeted surveillance). In either case, the working model of representation used by these systems consists of flawed shorthands that reveal a lack of critical thinking about what identity actually \textit{is}.

Inherited from a post-Cartesian empiricism that remains enshrined in much of scientific research, the default paradigm in AI research holds certain platitudes to be true about human identity: that it is composed of fixed, essential attributes. The basic tenets of statistical machine learning require either pre-existing categories of labeled data (in supervised learning) or seek to create such partitions within its training datasets (in unsupervised learning). In translating identity to data, arbitrary facets are sorted into \textit{discrete}, often mutually exclusive categories. Such models tacitly assume that identity is fixed over contexts: \textit{static} in its totality. They hold that identity is \textit{essential}—that there is some intrinsic, self-encompassing, ground-truth quality to it. These conceptions neaten the problem of representation mathematically, but fall short in faithfully reflecting identity as it operates interpersonally. Various components of identity are flattened when a person is represented by a system. 

Of course, technological abstractions are just that—abstractions— but we argue that the simplifications used by the vast majority of AI practitioners to capture identity directly impede their utility, and perpetuate harm on both the individual and societal level by immobilizing existing norms. Some abstraction is necessary for a model to be useful, but this particular flattening of identity strips away what \textit{is} useful about it. Among humans, identity can be used: as grounds for mutual understanding, as performance, as signals to an in-group for acceptance, as labels for dominance and subjugation, as recognition of self amid the crowd. Above all, identity exists as relations that are \textit{constantly in flux}. To a machine, human identity is useful insofar as it can figure into a working model of the problem. Not only are these existing abstractions inaccurate and thus limited in their predictive capacity, these models also influence societal conceptions of identity and have the capacity to harm. Poor modeling weakens the ability of an algorithm deployed in social contexts and causes it to make false sense of noise. Whatever predictions it outputs will make prescriptions upon individuals based on erroneous signals that perpetuate existing inequalities and manufacture new ones. As such models are deployed at scale today, inadvertent harms metastasize within feedback loops of human and machine.

Before further analysis, we must situate artificial intelligence research and application as it exists within late capitalism. Most impactful research today is born out of inextricable ties between government, industry, and academia, primarily within the Global North. Inevitably downstream from research comes the application. Trace any research purpose out and sooner or later you will feel the pulse of capital: market and shareholder demands placed on private corporations, nation-states engaging in imperial struggles within our globalized economy. This makes the relations between the algorithms deployed today and ourselves an asymmetrical one. Call it techno-feudalism \citep{morozov2022critique} or surveillance capitalism \citep{zuboff2019age}, but by any analysis, the interaction is ultimately extractive. While a full study of these systems is beyond the scope of this paper, it bears noting that algorithmic interactions with identity exist for a financial purpose, whether it is to create fixed subjects, mine data, or capture attention. This calls into question whether it is truly beneficial for the machines of today to have a better grasp of identity, given their makers. We suggest that representing identity more fluidly is not necessarily synonymous with higher resolution data; structural precision does not necessarily mean more detail to be extracted. Rather, such representation to a machine can provide us with more agency than the existing model. Perhaps it can even allow us to become evasive, unprofitable subjects. 

Strands of thought in critical theory and interrelated fields propose a far more dynamic understanding of identity than the simplifications made within AI research. A firm definition of identity is evasive, but theorizing about it as a \textit{system of relational processes} is key. Incorporating these concepts into machine learning frameworks is necessary not only in terms of fidelity to reality, but also in building systems that do not freeze existing norms. In Section \ref{autopoiesis}, we describe a theory of identity as what we call an “autopoietic system,” evolving processes of construction and function. This will be substantiated in Section \ref{theory}, which pulls theories of identity from fields of critical inquiry that challenge existing conceptions within AI research. We emphasize the importance of holding contradictions around identity, examine the indeterminacy of its materiality, and conceptualize it as an ongoing series of iterative events. In Section \ref{limitations}, we critique existing AI fairness approaches that fail to incorporate this ontology by perpetuating a discrete, static, and essential notion of identity. We emphasize the risks that ensue from this type of abstraction, using the following three dichotomies to characterize identity and structure our analysis.

\begin{enumerate}
  \item \textsc{discrete} vs. \textsc{continuous}
  \item \textsc{static} vs. \textsc{contextual}
  \item \textsc{essential} vs. \textsc{co-constructed}
\end{enumerate}
Finally, Section \ref{alternatives} imagines what an alternative framework that incorporates these qualities looks like. We outline two technical approaches to model design involving multilevel optimization and relational learning, then sketch an imaginary of what better systems could do. What possibilities are available to us with models that are capable of representing a fluid system of human social identity; what biases are circumvented; what avenues of feedback emerge afresh between human and machine? 

By adjusting our collective grasp on what ``identity'' consists of as a field, and relinquishing poor abstractions in favor of more precise ones, we hope to mitigate harms caused by fundamental misconceptions within the frameworks that underpin the technologies we deploy. The effects of AI reverberate through society en masse; it is up to us what they will look like.

\section{Identity as Autopoiesis}
\label{autopoiesis}

What do we mean when we talk about ``identity''? Do we conceptualize it as a fixed substance interpolated through the world, or rather a discursive narrative we tell about ourselves and others? Identity has emerged as a shorthand for a cache of attributes to apply to people (race, class, gender, etcetera), but is best understood as an action: to identify. You identify \textit{with}, or disidentify, using relational terms to situate yourself among the categorized crowd. Reconstrued as an action, we can then ask: identification with \textit{what}? Identification operates in the realm of the imaginary, as a process of differentiation that involves superimposing layers of fantasy over the identified subject. Your fantasy, their fantasy, your fantasy of their fantasy, their fantasy of yours. This is a non-deterministic, context-sensitive happening which nevertheless sediments into a sorting pattern, a complex classification, a “necessary negotiation between detail and abstraction” \cite{bowker2000sorting}. Our collective categorizations become a nascent topography, invisible and in flux, yet still with potent ramifications.

It is easy to suppose that underneath these fantasy layers there is some substance to which they are hooked, that the \emph{what} of identification can be unearthed. But this \textit{what} is indeterminate in the same way the process of identifying is: there is no fixed quality of a person that is constant across \textit{all} identifications made by \textit{all} people across \textit{all} time. What remains in dialogue is the overlap of imaginaries within identification processes. On a collective scale, these coagulate into the social norms used for identifying, which Judith Butler calls “cultural intelligibility” in \textit{Gender Trouble} \cite{butler2003gender}. What we recognise as identity’s substance is itself a product of unfurling fantasies of what identity might consist of. The recursion is readily apparent when any perceived substance of identity is placed under close scrutiny. While identity’s substance is often illusory, its downstream impact is undeniably material. Identity is integral to the human experience, philosophically and psychologically. Its relevance to politics rests in its irreducible existence as a vector of power: how it mediates relationships between subjects. All perceived differences are capable of generating exclusion and power differentials, substantiated or otherwise. 

In 1972, biologists Humberto Maturana and Francisco Varela introduced the term ``autopoiesis'' to describe a network of processes that is capable of reproducing and maintaining itself, in order to capture the self-contained chemistry of living cells \cite{maturana1991autopoiesis}. Such a system would continuously regenerate its own components and organization, realizing the very processes which produced them. By conceptualizing identity as an autopoietic system comprising its \textit{construction} and \textit{function}, it is possible to foreground the self-reinforcing circularity and perpetual motion of its processes. \textit{Construction}, therefore, refers to the processes by which identity is formed, discursively and psychically. \textit{Function} refers to the processes through which identity is used, internally and interpersonally. Each set of processes informs the other, constituting the other’s parts; the two can be co-located within the same social interaction. This abstraction allows for the contradictions inherent to defining identity. Feminist and queer theory is contentious over how the body figures into identity, but this ontology makes no claim to locate from where precisely the construction or function is derived. Instead, it describes the processes by which the concept takes form.

    \begin{figure}[h]
    \centering
    \includegraphics[scale=0.8]{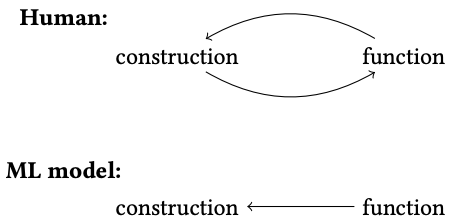}
        
    \Description{A graph of how identity operates for humans, with an arrow from function to construction, and an arrow from construction to function; next to a graph of how identity operates for machines, with only an arrow from function to construction.}
    \caption{Diagrams for the identity processes of humans (bidirectional) vs. machines (unidirectional).}
    \label{fig:system_comparisons}
    \end{figure}

Visualizing identity as an autopoietic system of interaction reveals how it is capable of evolving. As it is constituted of feedback loops, perturbing any norm around either how identity functions or is constructed can cascade across the entire network. Social drift is possible. The machine learning abstraction of identity, however, is not that of an autopoietic system, but rather a unidirectional one: how human identity is construed is derived purely from its utility to the algorithm (Fig. \ref{fig:system_comparisons}). This immobilizes the processes by which identity evolves in society as weak AI becomes a ubiquitous interlocutor in everyday life. Every existing category is reified. Without critically reframing identity in the field, hierarchies of power will calcify in the same way.

\section{Theorizing About Identity}
\label{theory}
Theorizing identity as feedback loops of construction and function makes plain the inaccuracy of how it is presently thought of in the realm of AI. Before critiquing machine learning approaches in Section \ref{limitations}, it is important to substantiate and clarify the claims made around this ontology. In this section, we corroborate the concept of identity as an autopoietic system by drawing from an inexhaustive list of theorists such as Butler, Stuart Hall, Karen Barad, and Jasbir K. Puar. We foreground the contradictions inherent to approaching identity, discuss how to understand its indeterminate materiality, and map out the cyclical, iterative nature of its processes.

\subsection{Holding contradictions}

Contradictions are inherent to theorizing around identity. They are what make it so difficult to contend with. In his introduction to the essay collection \textit{Questions of Cultural Identity}, Stuart Hall asks what accounts for identity’s continued relevance across disciplines, even as it has been critiqued and deconstructed for decades. He observes that the deconstructive approach theorists have taken to identity does not \textit{replace} its key concepts with “truer” ones but simply marks them as no longer serviceable. Yet by not superseding them with positive concepts, we are left with no option but to continue to use them in their deconstructed forms. That is, “the line which cancels them, paradoxically, permits them to go on being read” \cite{hall1996questions}. The concept of identity exists in a liminal space between “reversal and emergence,” what Hall describes as “an idea which cannot be thought in the old way, but without which certain key questions cannot be thought at all.” To critique the social order we are subsumed in, we must necessarily engage in this compromise.

Feminist theory has grappled with this paradox, in which the hermeneutics of suspicion are cast upon the categories that define their subjects and their tools of liberation. In \textit{Gender Trouble}, Butler observes that the assumption that “woman” refers to a common identity is anxiously contested, yet remarkably difficult to displace \cite{butler2003gender}. Gender cannot be a stable signifier, because it is impossible to parse it from the context in which it is generated. But there is a concurrent desire to presume a unitary subject of feminism as a basis for solidarity. Butler claims this is antithetical to the aims of feminism, and asks: “Is the construction of the category of women as a coherent and stable subject an unwitting regulation and reification of gender relations?” Indeed, the usefulness of such a category is precluded by the normative prescriptions it makes and the negative foreclosure implied by any variation of its definition. But categorization is inevitable, and we might ask, if there is no subject to fix, how can we cohere the discrimination that nevertheless occurs? There is no neutral position outside of the “contemporary field of power,” no refusal of that which is thrust upon you. Butler suggests that a discrete category actually undercuts any claim to representation, and will only be possible “when the subject of `women' is nowhere presumed.” Only by deferring any closure around identity categories are we able to critique the normative structures that immobilize them.

There are also necessary contradictions required to hold identity together within oneself, as a subject. Hall describes identification as a process of recognition, lodged in some common origin or shared attribute, a signifying practice that is ultimately always conditional. Identity is grounded in projection and idealization, and it is necessary to accept that “identities are never unified and, in late modern times, increasingly fragmented and fractured; never singular but multiply constructed across different, often intersecting and antagonistic, discourses, practices and positions.” He holds that identities only function through difference, “\textit{because} of their capacity to exclude, to leave out, to render `outside', abjected.” To Hall, the task at hand is then to theorize:

\begin{quotation}
[...] what the mechanisms are by which individuals as subjects identify (or do not identify) with the `positions’ to which they are summoned; as well as how they fashion, stylize, produce and `perform’ these positions, and why they never do so completely, for once and all time, and some never do, or are in a constant, agonistic process of struggling with, resisting, negotiating, and accommodating the normative or regulative rules with which they confront and regulate themselves.
\end{quotation}
Identity thus requires us to articulate the “point of suture” between the discourses and practices which fix us as subjects, and the processes which allow us to speak. We attach ourselves to discursive concepts and signifiers, while holding the understanding that they are always inadequate representations. This constellation of practices of “self-constitution, recognition, and reflection” produces the identified self within the world.

\subsection{Materiality and indeterminacy}
Before discussing the characteristics of processes which constitute identity, we must situate the material within its system. It is tempting to imagine there is a singular substance, some indisputable grounds of “oneness” on which to assign bias to and leverage against, forming the categorical boundary of any given identity. This supposed trait can be material or discursive (think of the urgency for feminists to articulate some universal structure of patriarchal domination) but we can imagine the imperative need for one to exist. Regardless, Butler rejects this rhetoric and argues that such cohesion is antithetical to the liberatory desires of any group, instead forming a closure that reifies constraint. What remains as a tool is not an essentialist but rather a strategic concept of identity. 

But is there some materiality at play within the system? Certainly—social interaction produces material effects which propagate onward, including into identity formation. Hall argues that there is no stable core of the self, but rather a continuous process of \textit{becoming}. Meaning, there is no true ``essence'' of self hiding within the superimposed selves held in common between those with a shared historicization. The question then, is that of: “using the resources of history, language and culture in the process of becoming rather than being: not `who we are' or `where we came from', so much as what we might become, how we have been represented and how that bears on how we might represent ourselves.” Precisely \emph{because} identity is partly imaginary and symbolic, constructed discursively, is why we must understand the routes in which it forms: the historical and political institutions through which it emerges. 
	
What Hall conceives of as “a process of articulation, a suturing, an over-determination not a subsumption,” can be extended in machine learning parlance to make sense of identification. This process of becoming is always an uncomfortable fit, and mediated through imperfect attachments to discursive fantasies. Identification, \emph{becoming}, is a relational process through which we attempt to parse and organize the world and situate ourselves in it. In other words, it is pattern recognition, an extrapolation of information, a coherence around working definitions we place and find put upon us. Categorization here is done ambivalently and with a neutral valence; there is nothing inherently harmful about this process. Categorization, the naming, the making visible, is what protects us from the “tyranny of structurelessness” \cite{freeman1972tyranny}. But every category reinforces one point of view, and erases another. In \textit{Sorting Things Out: Classification and its Consequences}, Geoffrey C. Bowker and Susan Leigh Star argue that to classify is human and such classifications are invisible and inescapable \cite{bowker2000sorting}. There are classifications formalized through institutions and bureaucracies, and ad hoc ones that are tacit in our minds. What is crucial to ask about them is: “Who makes them, and who may change them? When and why do they become visible? How do they spread? What, for instance, is the relationship among locally generated categories, tailored to the particular space of a bathroom cabinet, and the commodified, elaborate, expensive ones generated by medical diagnoses, government regulatory bodies, and pharmaceutical firms?” We extend this line of questioning to ask, what is the relationship between the malleable identity categorizations within our minds and those which are set out by a designed machine learning model? The task at hand is not to dispel categories entirely, but scrutinize the structures that create and maintain their forms. In Bowker and Star's words: “We need a topography of things such as the distribution of ambiguity; the fluid dynamics of how classification systems meet up—a plate tectonics rather than a static geology.”
	
In analyzing category systems, Bowker and Star also emphasize the importance of holding the past as indeterminate as a methodological practice. That is, knowledge of the past is under constant revision from the ``historical present.'' This indeterminacy is crucial to understanding the difference between human concepts around identity and those of an algorithm’s. Butler deconstructs the deterministic fiction of gender, which conceives of the body as a “passive medium” on which cultural concepts are inscribed. But bodies are not passive, and much of feminist theory is dedicated to showing that the frame of reference used to understand gender, the very \emph{language} by which it is critiqued, should be placed under suspicion. According to Butler: “As a shifting and contextual phenomenon, gender does not denote a substantive being, but a relative point of convergence among culturally and historically specific sets of relations” \cite{butler2003gender}. Any conception of what gender and other identities can be, is always relative to and inextricable from the mercurial processes through which it is determined. 
	
What indeterminate understanding of the materiality of identity can we then develop? How can we both incorporate and critique the matter through which processes of identity move? Karen Barad argues that fitting materiality into the discursive is vital in \textit{Posthumanist Performativity: Toward an Understanding of How Matter Comes to Matter} \cite{barad2003posthumanist}. They critique the overly representationalist approach taken by theorists to conceptualize discursive forces and offer an ontology of relations as a replacement. This “agential realist” framework resonates with our concept of identity as autopoiesis. It fundamentally construes phenomena as the primary ontological unit and stipulates that all entities \textit{only exist via interaction} (Barad uses the neologism \textit{intra-action} to imply there is no independent existing entity prior to such interaction). Within this graph of interaction, cuts can be temporarily made to enact a local resolution, some local causal structure, which allows one part of the universe to become “differentially intelligible” to another. The discursive practices which produce these cuts are thus specific acts of boundary-making, which permit intelligibility but nevertheless make no claim to the ongoing, dynamic whole. This framework holds both the multiplicity of matter and its indeterminacy. Machine learning algorithms produce such a cut within the network to make sense of identity, but make the tacit claim of establishing both a fixed material and a deterministic way of handling it. Instead of a universalizing project, such models should assume that their purpose in fact \textit{constructs} their underpinning materiality.

\subsection{Identity as event and iteration}
To finally clarify the lapse of machine learning when it comes to identity, we will cohere two characteristics that were gestured to already: identity as \textit{event} and, by extension, identity as \textit{iteration}. In the essay \textit{“I would rather be a cyborg than a goddess”: Becoming-Intersectional in Assemblage Theory}, Jasbir K. Puar suggests supplementing intersectionality with the concept of assemblage \cite{puar2012assemblage}. She is less concerned with the “formative, generative, and necessary intervention” of Kimberlé Crenshaw’s concept of intersectionality \cite{crenshaw1989demarginalizing}, but rather how effective invoking it is within the changing context of neoliberal mainstreaming of difference. “Much like the language of diversity, the language of intersectionality, its very invocation, it seems, largely substitutes for intersectional analysis itself.” It requires we produce a subject, to “infinitely multiply exclusion” to provide the appropriately specific, Othered one. Broadly pulling from scholars divested from subject formation, Puar offers the assemblage as an augmented framing to the grid produced by intersectionality.

Borrowed from the works of Gilles Deleuze and Félix Guattari, “assemblage” is a concept that more strictly emphasizes the \textit{assembling} of relations of patterns than the ensuing arrangement \cite{deleuze1988thousand}. Rather than producing a fixed constant against which to contrast variants, assemblages “foreground no constants but rather `variation to variation' and hence the event-ness of identity.” Analyzing identity as assemblage means paying particular attention to power and affect, while also blurring the boundaries of identity’s constituent parts. To illustrate, Puar rereads Crenshaw’s example of a crash at a traffic intersection, then Brian Massumi’s incident of domestic violence \cite{massumi2002parables} through the lens of assemblage:

\begin{quotation}
The difference between signification and significance (sense, value, force) is accentuated. There is a focus on the patterns of relations—not the entities themselves, but the patterns within which they are arranged with each other. The placements within the space itself have not necessarily altered, but the intensified relations have given new capacities to the entities. [...] There is a sense of potentiality, a becoming. “Anything could happen.” It is a moment of deterritorialization, a line of flight, something not available for immediate capture—“everything is up in the air,” and quite literally, the air is charged with possibility. Intersectional identity comes into play, as the (white) male is always already ideologically coded as more prone to violence. Finally, the strike happens: the hand against face. The line of flight is reterritorialized, forward into the social script, a closing off of one becoming, routed into another assemblage.
\end{quotation}
Re-envisioned as an event, identity \textit{includes} the forces and patterns of relations through which it takes form; the objects populating the scene, human and otherwise, are all in play, all bound in tight relation to one another. Identity as an event, as assemblage, de-emphasizes the subject and allows a sensitive analysis that can contend with the multiplicity of layered relations and their continuous movement through space-time.

As an extension, identity does not simply occur as an event; it occurs as an \textit{iterative} event. Butler clarifies their notion of performativity in \textit{Bodies That Matter}, the idea that gender is a “doing” \cite{butler2011bodies}. Rather than being equated to simply performance, performativity “cannot be understood outside of a process of iterability, a regularized and constrained repetition of norms. [...] This iterability implies that ‘performance’ is not a singular ‘act’ or event, but a ritualized production, a ritual reiterated under and through constraint.” The most critical element to highlight here is the recursive element, the immediate feedback loop. Performative identity always occurs with a response which restrains and modifies it. During the process of identification, the subject takes on certain prescriptions and aspires to others; Butler calls this aspect of performance ``citationality.'' Yet as Hall describes, identification is always a misfit: “an over-determination or a lack, but never a proper fit, a totality.” This iterative process both enforces and destabilizes the very identity concepts it interacts with.

Therein lies the fundamental distinction between how identity works among humans, and how it is handled by a machine learning model. Among humans, the subversion of identities, and indeed all conscious or unconscious enactments and interactions, means that any conceived “identity” is perpetually in flux; natural drift will and does occur. All emergent categories, as existing sediments of ritual, can be collectively eroded and take on new forms. The relations which constitute identity, which we have described as \textit{construction} and \textit{function}, are far more plentiful and complex than a simple reducibility to those terms. Nevertheless, such an abstraction reveals the critical lapse within machine learning. There, the strict referent form of “identity” is decreed by whatever purpose it offers to the model and nothing else (without even interrogating that purpose). Towards an algorithm, there is no closing of the loop back from \textit{construction} to \textit{function}, no avenue of recourse for any presumptions it places on identity. What discrete categories and shapes its training data uses to give form to identity cannot evolve or iterate or be subverted. The machine conceives of identity: in stasis, as deterministic, as an empirical truth-attribute of being. No footnote about the limitations or shorthands made to \emph{make the model work} can undo the harms they might propagate as AI becomes omnipresent in our lives. What does it mean when these interlocutors of our experience freeze the fluid dynamics of identity? What prescriptions will be put upon us with no possibility of change? What reification will happen not only to extant identity concepts, but also the power structures they engender? A total ontological shift with regards to identity is required within the realm of AI research. In the ensuing sections, we will make a critical pass over existing fairness techniques involving identity and offer a rough sketch of what this shift could look like.

\section{The Limitations of AI Fairness}
\label{limitations}

Rethinking identity as an autopoietic system of formation and function highlights its circular nature.
In comparison, machine learning systems enforce a unidirectional relation: categories are formed and fixed according to their utility to the model.
AI fairness researchers have presented well-intentioned technical solutions to mitigate bias and enforce equality for protected groups. However, their abstraction of identity forms a precarious foundation for fairer systems. 
In this section we critique the ontological assumptions around identity in AI fairness by providing examples violating our theory of relational identity. We organize our assessment by three dichotomies informed by the theory laid out above.

\subsection{Discrete vs. continuous}

Many AI fairness approaches assume that identity is \emph{discrete}, using binary gender \cite{cho2021crosslingual} or mutually exclusive racial categories \cite{larson2016compas}. This opposes a \emph{continuous} representation of identity, which we can imagine as floating point values.
Even in one system using continuous inputs, the predicted attribute (rate of violent crime) is binarized to 0 (30th percentile of crime rate) and 1 (70th percentile of crime rate), to collapse the problem to classification \cite{kearns2017gerrymandering}. Discretization means erasure when identity operates in a complex continuous space. Gender is not a binary, but rather a shifting discursive act. Multiracial people are not a single category, and lumping them together is harmful negligence. Furthermore, people in the same racial ``category'' often receive different levels of discrimination due to colorism. When it comes to sexuality, the Kinsey scale reflects sexual orientation as a spectrum, rather than a binary of hetero- and homosexual (further enriched by multi-dimensional systems such as the Klein Grid and the Storms Scale \cite{storms1980theories}).

This lack of gradation often occurs because of the limitations of the chosen model, such as when classification (which outputs discrete categories) is chosen for prediction instead of regression (which outputs real-valued predictions). In the case of the racial categories of Ionescu's study of race-based homogamy in \cite{ionescu2021homogamy}, race is reduced to black and white because the agent-based model utilized could only handle binary features. Even the Gender Shades study, which analyzed colorism in gender classification systems, chooses discrete categories for labelling skin tone rather than a continuous variable \cite{buolamwini2018gendershades}.
The decision to discretize identity can be made anytime during system design, including during dataset collection and curation, before model constraints come into play. The ProPublica COMPAS dataset, created to analyze racial bias in recidivism predictions, divides race into six categories: black, white, Hispanic, Asian, Native American and ``Other'' (a category describing 343 out of 80,000 defendants) \cite{larson2016compas}.

An argument for discretization is: people often make categorical judgments about social identity, so why should machines differ? Indeed, multiracial people are often treated as a single race at both an interpersonal and systemic level, called \emph{monoracial normativity} by a 2019 study illustrating the phenomenon \cite{ford2019monoracial}. But human beliefs are ultimately malleable, while the categories programmed into machine learning systems are not.
The risk of discrete categories emerges when machine inference informs action; after all, models are deployed so they can be used. A low-dimensional discrete prediction can only produce simplified actions that may not be proportional in context. In the case of binarized violent crime risk rates, if predictions decide whether extra police are deployed to a neighborhood, those on the precipice of the threshold will be particularly ill-served and experience a higher rate of false arrests and police violence. Deploying any predictive policing heuristic forms a feedback loop that reinforces carceral injustice, but this illustrates additional harms caused by discrete classification.

Using real-valued variables solves some of these problems by allowing gradations of difference. However, they do not fully satisfy the autopoietic model of identity. Continuous values do not capture the dynamism of identity.
Even placed on a continuum, our skin color does not determine ``how much'' we identify with a racial category. Upbringing and cultural context informs our relationship with race and its perception (consider adoptees, immigrants, multiracial people). We code-switch, linguistically and behaviorally, based on whether we are at work, at home, among friends, walking down the street. Indeed, this phenomenon is present for many facets of identity.
What equation can possibly capture the social complexities of ``how Black'' someone is, or ``how much of a woman''? The answers to these questions, insofar as they can be answered, differ across contexts.

\subsection{Static vs. contextual}

A \emph{static} identity is immutable across time and unchanged by other variables in the data, observable or otherwise. In contrast, understanding identity as \emph{contextual} recognizes how it changes across time and space. To contrast the continuous conception of identity with the contextual, consider a fictional example. Alice is labeled as ``heterosexual,'' but actually identifies as ``mostly straight'': this continuous trait is discretized in most classifications. Meanwhile, Alice's colleague Bob records his identity as heterosexual on a workplace survey while seeking out only same-sex relationships on a dating app, because he prefers not to disclose his sexuality at the office. Now, imagine Bob's employer-sponsored insurance company deploys a ML system that ingests a massive amount of anonymized employer-gathered data as well as an employee's personal web browser history. Maybe the designers of the system want to detect the sexual orientation of employees because this is a protected characteristic, and they want to ensure fair outcomes for people like Bob. Are there frameworks in AI fairness that account for Bob's identity changing in context?

While some tools in machine learning handle context and mutability, identity labels are typically made static in AI fairness. In the influential frameworks of individual and group fairness, and offshoots such as subgroup and intersectional fairness, protected characteristics are considered static for all individuals.
As a solution, counterfactual fairness has been proposed for incorporating causal context into fairness analysis.
In this type of analysis, counterfactual reasoning can infer the influence of observable factors on other variables in the dataset. By constructing a directed acyclic graph of latent variables that influence observable variables, the inference process identifies causes via intervention: the substitution of different values for the latent variables \cite{kusner2017counterfactual}.

However, there are numerous critiques of how counterfactual fairness fails to accurately reflect identity dynamics. Kohler-Hausman argues that using the causal inference framework to detect racism reduces race entirely to phenotype, neglecting the interplay of internal experience and societal sculpting \cite{kohler2018eddie}.
Kasirzadeh et al. \cite{kasirzadeh2021counterfactuals} offers a similar critique and advises systems designers to consider the semantic and ontological origins of identity categories in causal graphs.
In \cite{hu2020sex}, Hu recognizes the ontological problem of representing social categories as nodes on the causal graph, instead of regions in a dense, ever-shifting web of relations. These are corollaries to our theory of identity as an autopoetic system. 
We launch a complementary critique by acknowledging contextual identity. Counterfactual fairness does not allow identity to vary contextually, but instead compares individuals with varying sets of characteristics as counterfactual evidence for each other's outcomes. In this approach, interventions are made by generating fictional individuals with counterfactual identities. However, in a model of counterfactual fairness, the \emph{same} individual would display different identity characteristics under different circumstances. This subtle difference has implications for fair outcomes. In fact, Kusner emphasizes in the supplemental of \textit{Counterfactual Fairness} that the influence of protected attributes on decisions relies upon interventions \emph{across} individuals, not \emph{within} the same individual.

Beyond this, handling context does not complete a system's view of identity. Context-sensitivity alone permits labels that are mutable but deterministic, in which a fixed set of circumstances can change the contextual variable. But a system that does not constantly revise its cycle of construction and function fails to fulfill our theory of identity processes. Epistemological questions are left unanswered. Who determines what entities and signifiers are contained in a dataset? How are the signs that comprise a model's conception of identity assembled? What are the dangers of assuming a ground truth?

\subsection{Essential vs. co-constructed}

Machine learning systems tacitly assume that the signifiers within its data reflect inherent properties of the entities represented. Predictions, inferences, and calculations operate within the rigid semiotics set by the design of the dataset and the model. When they include identity labels, the system interprets the classification of people as a knowable, objective, \emph{essential} truth.
This contrasts the theory of \emph{co-construction}, which frames identity as an ongoing interaction, generated indeterminately. While discrete simplification begs the question, ``Can one's identity exist on a spectrum?'' and static simplification begs, ``Can one's identity change over time and space?'' the essential simplification escalates: ``What is actually represented by one's categorical signifiers, and from whose point of view?''

Essentialism assumes identity is an innate property capturing the subject's ``essence.'' We can connect essentialism in machine learning to a colonial scientific tradition running through inequitable medical treatment \cite{obermeyer2019dissecting}, eugenics \cite{cave2020problem} and the reinforcement of colonial power and norms \cite{mohamed2020decolonial}.
In \emph{Truth from the machine}, Keyes et al. present a case study of how the assumption of inborn causes for autism and homosexuality led to faulty conclusions in data science-driven studies \cite{keyes2021materialization}.
Sociotechnical systems contain categories within their structure, and thus any system attempting to enact fairness along axes of identity necessarily reinforces their boundaries. Such a system may introduce harms to invisibly marginalized groups (though invisibility can even be desirable). Such systems also have long-term implications, by perpetuating a static set of standards while social norms drift. Debiasing machine learning systems with respect to protected characteristics is a significant goal for AI fairness research, but if protected categories are drawn along essential lines, these models risk reinforcing systemic inequality.

Recent works interrogate the epistemic basis of identity representations in AI.
Hanna, Denton, et al. \cite{hanna2020towards} set out guidelines for applying critical race theory to understand algorithmic categorization of race and ethnicity.
Denton additionally describes a genealogy of AI in order to confront the histories and norms embedded in datasets \cite{denton2020bringing}.
The practice of ``studying up'' in machine learning reverses essential assumptions by turning the lens of machine predictions on the social norms and cultural context of those holding power \cite{barabas2020studying}.
In a similar vein to our work, Hancox-Li et al. critiqued the limitations of feature importance methods and suggests methodologies from feminist epistemology to address them \cite{hancox2021epistemic}. In their writing, context sensitivity and interactive ways of knowing agree with our theory of identity as ongoing relational processes.

While the AI fairness space is actively engaged in the critique of essential identities in supervised and unsupervised learning, identity in reinforcement learning is underexplored. 
Even so, essentialism is already emerging. Developing human-compatible AI is an increasingly prominent goal in RL, aimed at training agents to cooperatively interact with humans. Several recent projects aim to improve compatibility through a ``type’’ framework, in which agents are constructed to model the type of other agents in their environment. Here, ``type'' represents some abstract feature parameterizing the entity`s behavior (e.g., their policy or action distribution). The aim of modeling types is to improve the model's on-the-fly adaption to new individuals—including humans. However, these projects do not interrogate the ontology of a ``type.'' For example, Ghosh et al. \cite{ghosh2020towards} train reinforcement learning agents to infer the ``true type’’ of their partner in a cooperative game. The concept of a true type may be sensible for artificial agents, since they can be encoded with particular properties. As noted above, though, human identity has no ``true’’ value—and so an approach based on such types is intrinsically fraught.
Nikolaidis et al. \cite{nikolaidis2015efficient} adopt a similar type-based approach to work on human-robot collaboration. The authors use an unsupervised learning algorithm to cluster behavioral demonstrations into different types, arguing that a ``limited number of `dominant’ strategies” can account for the majority of demonstrations. The project introduces human types as a partially observable variable in a Markov decision process, allowing improved collaboration for agents at test time. Again, the concept of type leaves human feedback out of the model’s construction of identity. On one hand, type-based approach may be able to help support inclusivity, by generalizing to a diverse population of individuals. However, it also risks setting an essentialist precedent in the expanding field of reinforcement learning with human-agent interaction. 

The co-constructivist stance holds semantic categories as fluid and subjective. For any given identity category, people have their own adaptable notion of its semantic meaning, regardless of their relation to it. It is therefore impossible—or at least ill-advised—to build a machine learning system that discovers, classifies or infers ``objective'' identity labels, or draws ``objectively correct'' conclusions from identity data. Prior works either elide this point or leverage it to critique essentialist paradigms in science and technology without proposing alternatives. In contrast, the next section takes a positive position. Humans make subjective classifications all the time. We constantly update our abstractions; we hold contradictory viewpoints; we imagine the experiences and preferences of others. Ultimately, our conceptions of identity are fluid and susceptible to change, play, and subversion. Is it possible to change conceptions of identity within machine learning models to be the same?

\section{Alternative System Configurations}
\label{alternatives}
While the field of AI fairness offers a growing number of technical solutions, these often operate under the same assumptions as the models they criticize. The majority of solutions do not allow the circular, iterative nature of identity; rather, they continue to assume it is composed of fixed, essential attributes. They immobilize identity concepts and do not permit the possibility of drift or subversion. Broader critiques of machine learning fundamentals are more aligned with this paper's theory of identity as interaction, but they do not venture positive solutions, and some even claim the discipline itself is fundamentally at odds with the ambiguity of human behavior \cite{birhane2021ambiguity}. We acknowledge with full modesty that machine learning systems have epistemic limits when it comes to understanding of identity (just as we do). However, within these limits exists a rich space of system configurations that are underexplored in the applications of AI fairness. Conceptualizing identity as autopoiesis, comprising processes of \textit{construction} and \textit{function}, also permits new imaginaries about how interaction \textit{between} identity systems and machine learning systems can play out: intra-system interplay. We offer two provocations to practitioners contending with AI fairness and identity:

\begin{enumerate}
\item Within machine learning systems, how can we close the loop from identity \textit{construction} to its \textit{function}? How can machines form internal concepts of identity that allow: mutability, iteration, social drift?
\item What interplay is possible when such machine learning systems are integrated \textit{into} identity systems? What new forms of co-construction, destabilization, and subversion are possible with the machine as interlocutor?
\end{enumerate}
We hope these open questions stimulate new avenues of research, and map out possible directions by offering concrete technical frameworks and loose theoretical imaginaries. To close the loop within machine conceptions of identity, we sketch two complementary approaches inspired by the theory presented in Section \ref{theory}: multilevel optimization and relational learning. To imagine new forms of identity co-construction between human and machine, we finish by illustrating scenarios of intra-system interplay, when machines \textit{can} conceive of fluctuating identity. 

\subsection{Technical approaches to closing the loop}

\subsubsection{Autopoiesis as multilevel optimization}

Society collectively constructs a concept of identity to suit certain functions, but individuals also define necessary functions based on their identity. Can we capture this cyclic interplay of construction and function with a computational model? A possible tool for describing our model of identity is \emph{bilevel optimization}, in which one model is optimized with respect to the optimum of another:

\begin{align*}
x^* &= \argmin_{x \in X}{F(x, y^*(x))} \\
y^*(x) &= \argmin_{y \in Y}{f(x, y)} 
\end{align*}

The family of bilevel optimization encompasses two popular machine learning frameworks: generative adversarial networks (GANs) and actor-critic methods in reinforcement learning \cite{pfau2016multilevel}.
We separately consider these frameworks to understand if they suit the case of learning identity, then sketch the schematics of an identity learning system inspired by these techniques.

In the unsupervised GAN setting, the goal is to learn how to generate likely samples from the distribution of the input dataset. A generator network produces candidate samples to fool the discriminator, while a discriminator network classifies samples as real or generated.
An obvious application to identity is to train the discriminator to classify (or regress) samples to an identity label, and optimize the generator to produce samples that defy categorization. These adversarial identity examples could qualitatively illustrate characteristics that resist implicit norms. This approach also expresses the contextual, fluid property of our model of autopoietic identity. Identity boundaries shift over time as the generator and the discriminator interact. However, this setting requires supervision in the form of identity labels for people represented in the training data. Thus, this approach fails to overturn the default representation of identity in machine learning. In the final artifact of training, a contextual understanding of identity is not guaranteed. The features that constitute an identity category are essentially and unidirectionally determined by the dataset. Perhaps a different extension of GANs can more closely fit our model of identity, but we leave that exploration for future work.

In the actor-critic setting, the goal is to train an agent to output a sequence of actions that maximizes reward over time in its environment. A policy network produces actions that maximize the value function, which is learned by the critic function. Here, value is defined as the discounted sum of expected reward of a state/action pair (contingent on the policy, since it is the source of future actions). In contrast to the adversarial setting, the two objectives are not opposed, but their circular dependence can make optimization difficult in practice \cite{grondman2012survey}.
Imagine a Markov Decision Process (MDP) where the states are observable characteristics of a person over time and the observed reactions of others, while actions represent the person's identity labels in that state.
In this setting, policy learning is analogous to the individual's construction of their identity: what they choose to identify with over time, shaped by perceptual inputs, conferral of others, and individual preference. The critic represents the utility of the person's active identity in context. The actor-critic framework is not limited to discrete or continuous representations. The sequential nature of MDPs expresses the mutability of identity over time, which is missing from the GAN approach. Additionally, the actor and critic mutually construct features that constitute an identity from external and internal feedback.

What does the reward function represent in this scenario? One possible interpretation is that it is the individual's satisfaction with the identity label provided by the policy in the current context. Nonetheless, issues arise from this approach. Is satisfaction actively measured, or modeled? If humans are involved, how are they incorporated into the training loop? How will the system respond to identities not encountered during training? 
In an alternative configuration, the reward instead represents satisfaction within a specific task. 
Identification then becomes an auxiliary task that may or may not prove relevant, depending on how the human's preferences interact with the environment.
The policy associates identity actions based on their utility for the task at hand, according to the human's preferences, the responses of humans, or even the responses of agents. 
This proposal for actor-critic methods reiterates the importance of considering the potential uses and misuses of RL in AI fairness.

In contrast to many of the systems discussed above, this framework avoids the explicit priors that bias and immobilize identity. Nonetheless, it does not resolve epistemic issues in the dataset collection process, the choice of categorizations and representations, and the subjective positions within identity. In addition, it poses identity a function of the \emph{environment}, which is ill-defined and may elude a useful operationalization. An environment that enables the policy to learn identification must include both the preferences of the individual described and their surrounding social network. Consequently, in the next design sketch, we turn our focus to the relational nature of identity.

\subsubsection{Relational and subjective learning}

Hegemonic forces shape machine learning and are reflected in the systemic bias entrenched in standard datasets and benchmarks. For a case study of the power dynamics at play in the dataset collection process, see Miceli et al. \cite{miceli2020between}. As an alternative to Western universalism, Birhane proposes relational ethics as the guiding principle for more equitable sociotechnical systems \cite{birhane2021algorithmic}, which center the interconnectedness of all entities and define personhood in terms of an individual's relationships to others. Relational schools of thought include Afrofeminist philosophy, the Zulu tradition of Ubuntu, and Eastern traditions such as Daoism. It also motivates an ecofeminist approach by respecting the interdependence of humanity and nature.
\cite{mhlambi2020rationality} explores core concepts from Ubuntu to form a guiding framework for AI governance.
However, relationality has not seen much adoption in AI fairness systems design.

The autopoietic model of identity is fundamentally relational: it situates the individual's identity within a shifting network of interaction. Each individual in a social network will perceive another's identity differently based on their individual experience. What would a relational approach to \emph{learning} identity look like?
One might begin with a subjective dataset of identifications.
We can ask individuals in a community of interest to first describe their own identity, then describe the identities of the other participants, based on their subjective internal logics.
Such a dataset could train a wholly relational model of identity—subjective by construction, but reflective of the relations each individual brings to the collective identity system. We can condition a supervised classifier of identity on aspects of the beholder's self-identification (or relevant features of a latent identity embedding space, depending on the representation). In settings that also include data relevant to a resource distribution or allocation tasks, we can apply algorithmic fairness techniques to compute a set of subjectively fair outcomes, according to the different perceptions of identity and protected characteristics of individuals within the dataset. This operationalization of relations may be generally useful for capturing a network of semantics, where meaning is subjective and classification is contextual—from interpretations of the law and ethical codes, to biological taxonomies of organisms.

Of course, by definition this model will only generalize to the demographics of the pool of participants. As stressed by Bowker and Star \cite{bowker2000sorting}, all standards reinforce and erase; there is no ``view from nowhere.'' If certain groups are underrepresented with regards to the purpose and locale where the model will be deployed, then the quality of the conditional model will necessarily deteriorate.
The model will not extend past the perceptions and biases of those groups which form its training data.

The format of the data is also a key consideration. Rather than the typical predetermined vector with constrained dimensions containing options for race, gender, etc. we propose recording identity as a freeform text field with an open prompt. Training a language model on such a dataset, if it were sufficiently large, could yield a rich, varied latent representation of identity, and a conditional model for identity generation.
The externally perceived characteristics of individuals are open to customization based on the application of the system. Examples inspired by other fairness applications include facial photographs, resumes or job applications, healthcare records (noting that privacy should be carefully considered in any human dataset). Returning to Hall \citep{hall1996questions}, any observable sign of identity is ``never a proper fit,'' always an incomplete snapshot of a shifting landscape. The proposed dataset would offer many snapshots of the landscape formed by its subjects, and the role of the algorithm would be to stitch those snapshots into a map of relational identity.

\subsection{Theoretical imaginary of intra-system interplay}

Beyond technical frameworks that close the gap in identity formation, we sketch an imaginary of what we call ``intra-system interplay,'' where the systems in question are machine learning systems and autopoietic identity systems. Section \ref{autopoiesis} described how identity exists as self-contained loops of processes, roughly reducible to those of construction and function. This overarching set of relations encompasses our interactions with other people, but also everything else. In other words, identities are shaped not only by the people around us, but by the \textit{things} as well. Technology in the broadest possible sense can refer to a pencil, a wheel, a cellphone, or a machine learning model. It follows, perhaps obviously, that all the objects we are situated among, all the technologies we mediate our experiences through and with, figure into our identity system \cite{sterelny2010minds}. The crucial difference between machine learning systems with ameliorated understandings of identity and those other technologies, though, is that the ML models are also self-contained, ongoing systems. That is, they can \textit{go on producing identity without us}. 

To clarify: when a machine is capable of fully realizing identity in all its malleable, iterative, circular, relational glory, it holds that such a machine can also destabilize and expand identity concepts. What would that cause at a societal scale? What possibilities are unleashed when machine learning systems are no longer adversarial enforcers of identity categories? What would an assemblage of machine co-constructed, cyber-identities look like? Answers might coalesce already around online communities and curation algorithms that invoke new forms of human experience \cite{bivens2016baking}, but this line of questioning goes a step further. If/when our models can think of identity the way we do, we will have machines that can reveal new methods of subversion, reshape identity “configurations,” reframe \textit{our} fundamental conception of identity. This may sound like pure speculative fiction, but don't forget: identity is partly fiction too.

\section{Conclusion}

A useful concept of “identity” eludes machine learning practitioners, who are often caught between simplifying abstractions and stifling complexity. Theorizing about identity is mired with contradictions and indeterminacy, but framing identity as continuous iterative events make it legible. Rather than attempt to put forth a strict specification, we conceptualize identity as an autopoietic system. We understand it as processes of construction and function, that are cyclic and self-reinforcing yet pliable and amenable to subversion; this conception provides a bedrock for critiquing existing paradigms in AI and imagining new possibilities. Common assumptions made in machine learning sever the bidirectional relations within this network by fully configuring identity according to utility, and not allowing the opposite causal flow. AI fairness techniques meant to ameliorate identity-based harms often invoke the same assumptions: that identity consists of discrete, static, and essential attributes. We argue these practices erase the elements that evolve identity concepts, preventing social drift, subversion, and the possibility of open-ended reinvention.

Identity may be part-situated in fantasy, but this does not denigrate its impact or validity. In Section \ref{alternatives}, we offer two provocations, asking how we can close the loop \textit{within} machine learning systems' conception of identity, and what futures are available when machines are playmates in identity systems rather than adversaries. We outline a high-level schematic for possible systems with multilevel optimization and relational learning, and sketch out a new imaginary for human-machine identity formation. We encourage all machine learning practitioners to engage with these open questions. As artificial intelligence becomes ubiquitous in our lives, it plays a burgeoning role in our identity relations. Given this context, we call for a fundamental re-imagining of how we configure our machines. It is up to us whether we make machines that calcify existing identities and concurrent power hierarchies, or machines that help us expand the dimensions of mercurial possibility. 

\begin{anonsuppress}
\section{acknowledgements}
We thank Boxi Wu for suggesting source texts and early input, Guy Mackinnon-Little for theoretical contributions and rigorous editing, William Isaac for high-level guidance, Laura Weidinger for review and feedback, and Shakir Mohamed for valuable insights and leadership.

\section{Funding Disclosure}
The authors declare no additional sources of funding. Colleagues at DeepMind participated in the review and approval of the manuscript. Aside from the authors, DeepMind had no role in the design; or preparation of the manuscript; or the decision to submit the manuscript for publication. The authors declare no other financial interests.

\end{anonsuppress}

\bibliographystyle{ACM-Reference-Format}
\bibliography{sample-base}

\end{document}